\documentclass[a4paper,orivec,runningheads]{llncs}
\usepackage{gensymb}
\usepackage{makeidx}
\usepackage{amsmath}
\usepackage{booktabs}
\usepackage{epstopdf, tabularx, multirow, siunitx}
\usepackage{wrapfig}
\usepackage{graphicx}
\usepackage{hyperref}

\begin{document}

\titlerunning{3D Convolutional Networks for Brain Parcellation}
\title{On the Compactness, Efficiency, and Representation of 3D
Convolutional Networks:
Brain Parcellation as a Pretext Task
}

\author{Wenqi Li\and Guotai Wang\and  Lucas Fidon\and  Sebastien Ourselin\and  M. Jorge Cardoso
\and Tom Vercauteren}
\authorrunning{Li et al.}
\institute{Translational Imaging Group, Centre for Medical Image Computing
(CMIC), University College London, London, UK\\
Wellcome/EPSRC Centre for Surgical and Interventional Science,
University College London, London, UK
}
\maketitle

\begin{abstract}
Deep convolutional neural networks are powerful tools for learning visual
representations from images.  However, designing efficient deep architectures
to analyse volumetric medical images remains challenging.  This work
investigates efficient and flexible elements of modern convolutional networks
such as dilated convolution and residual connection.  With these essential
building blocks, we propose a high-resolution, compact convolutional network
for volumetric image segmentation.  To illustrate its efficiency of learning 3D
representation from large-scale image data, the proposed network is validated
with the challenging task of parcellating 155 neuroanatomical structures from
brain MR images.  Our experiments show that the proposed network architecture
compares favourably with state-of-the-art volumetric segmentation networks
while being an order of magnitude more compact.  We consider the brain
parcellation task as a pretext task for volumetric image segmentation; our
trained network potentially provides a good starting point for transfer
learning.  Additionally, we show the feasibility of voxel-level uncertainty
estimation using a sampling approximation through dropout.
\end{abstract}

\section{Introduction}



Convolutional neural networks (CNNs) have been shown to be powerful tools for
learning visual representations from images.
They often consist of multiple layers of non-linear functions with a large
number of trainable parameters.  Hierarchical features can be obtained by
training the CNNs discriminatively.

In the medical image computing domain, recent years have seen a growing number
of applications using CNNs.  Although there have been recent advances in
tailoring CNNs to analyse volumetric images, most of the work to date studies
image representations in 2D.  While volumetric representations are more
informative, the number of voxels scales cubically with the size of the region
of interest.  This raises challenges of learning more complex visual patterns
as well as higher computational burden compared to the 2D cases.  While
developing compact and effective 3D network architectures is of significant
interest, designing 3D CNNs remains a challenging problem.

The goal of this paper is to design a high-resolution and compact network
architecture for the segmentation of fine structures in volumetric images.  For
this purpose, we study the simple and flexible elements of modern
convolutional networks, such as dilated convolution and residual connection.
Most of the existing network architectures follow a fully convolutional
downsample-upsample
pathway~\cite{Kamnitsas201761,dou2016automatic,merkow2016dense,cciccek20163d,milletari2016v,kleesiek2016deep}.
Low-level features with high spatial resolutions are first downsampled for
higher-level feature abstraction;  then the feature maps are upsampled to
achieve high-resolution segmentation.  In contrast to these, we propose a novel
3D architecture that incorporates high spatial resolution feature maps
throughout the layers, and can be trained with a wide range of receptive
fields.  We validate our network with the challenging task of automated brain
parcellation into $155$ structures from T1-weighted MR images.  We show that
the proposed network, with twenty times fewer parameters, achieves competitive
segmentation performance compared with state-of-the-art architectures.

A well-designed network could be trained with a large-scale dataset and enables
transfer learning to other image recognition tasks~\cite{huh2016makes}.  In the
field of computer vision, the well-known AlexNet and VGG net were trained on the
ImageNet dataset.  They provide general-purpose image representations that can
be adapted for a wide range of computer vision problems.  Given the large
amount of data and the complex visual patterns of the brain parcellation
problem, we consider it as a pretext task.  Our trained network is the first
step towards a general-purpose volumetric image representation.  It potentially
provides an initial model for transfer learning of other volumetric image
segmentation tasks.

The uncertainty of the segmentation is also important for indicating the
confidence and reliability of one
algorithm~\cite{Gal2016Dropout,sankaran2014real,shi2011multi}.
The high uncertainty of labelling can be a sign of an
unreliable classification.
In this work, we demonstrate the feasibility of voxel-level uncertainty
estimation using Monte Carlo samples of the proposed network with dropout at
test time.  Compared to the existing volumetric segmentation networks,
our compact network has fewer parameter interactions, and thus potentially
achieves better uncertainty estimates with fewer samples.

\section{On the elements of 3D convolutional networks}

%
%

\subsubsection{Convolutions and dilated convolutions.}
To maintain a relatively low number of parameters, we choose to use small 3D
convolutional kernels with only $3^3$ parameters for all convolutions.  This is
about the smallest kernel that can represent 3D features in all directions with
respect to the central voxel.  Although a convolutional kernel with $5\times
5\times 5$ voxels gives the same receptive field as stacking two layers of
$3\times 3\times 3$-voxel convolution, the latter has approximately $57\%$
fewer parameters.  Using smaller kernels implicitly imposes more
regularisation on the network while achieving the same receptive field.

To further enlarge the receptive field to capture large image contexts, most of
the existing volumetric segmentation networks downsample the intermediate
feature maps.  This significantly reduces the spatial resolution.  For example,
3D U-net~\cite{cciccek20163d} heavily employs $2\times 2\times 2$-voxel max
pooling with strides of two voxels in each dimension.  Each max pooling reduces
the feature responses of the previous layer to only $1/8$ of its spatial
resolution.  Upsampling layers, such as deconvolutions, are often used
subsequently to partially recover the high resolution of the input.  However,
adding deconvolution layers also introduces additional computational costs.

Recently, Chen et al.~\cite{chen2016deeplab} used dilated convolutions with
upsampled kernels for semantic image segmentation.  The advantages of dilated
convolutions are that the features can be computed with a high spatial
resolution, and the size of the receptive field can be enlarged arbitrarily.
Dilated convolutions can be used to produce accurate dense predictions and
detailed segmentation maps along object boundaries.

In contrast to the downsample-upsample pathway, we propose to adopt dilated
convolutions for volumetric image segmentation.  More specifically, the
convolutional kernels are upsampled with a dilation factor $r$.  For
$M$-channels of input feature maps $\mathbf{I}$, the output feature channel
$\mathbf{O}$ generated with dilated convolutions are:
\begin{align}
  \mathbf{O}_{x,y,z} =
  \sum_{m=0}^{M-1}\sum_{i=0}^{2}\sum_{j=0}^{2}\sum_{k=0}^{2}
  \mathbf{w}_{i,j,k,m}\mathbf{I}_{(x+ir),(y+jr),(z+kr),m}\,;
  \label{dilated_conv}
\end{align}
where the index tuple $(x,y,z)$ runs through every spatial location in the
volumes; the kernels $\mathbf{w}$ consist of $3^3\times M$ trainable
parameters.  The dilated convolution in Eq.~(\ref{dilated_conv}) has the
same number of trainable parameters as the standard $3\times 3\times 3$
convolution.  It preserves the spatial resolution and provides a
$(2r+1)^3$-voxel receptive field.  Setting $r$ to $1$ reduces the dilated
convolution to the standard $3\times3\times3$ convolution.
In practice, we implement 3D dilated convolutions with a split-and-merge
strategy~\cite{chen2016deeplab} to benefit from the existing GPU convolution
routines.

%
%


\subsubsection{Residual connections.}

\begin{wrapfigure}{r}{0.35\textwidth}
  \vspace{-25pt}
  \includegraphics[width=1.0\linewidth]{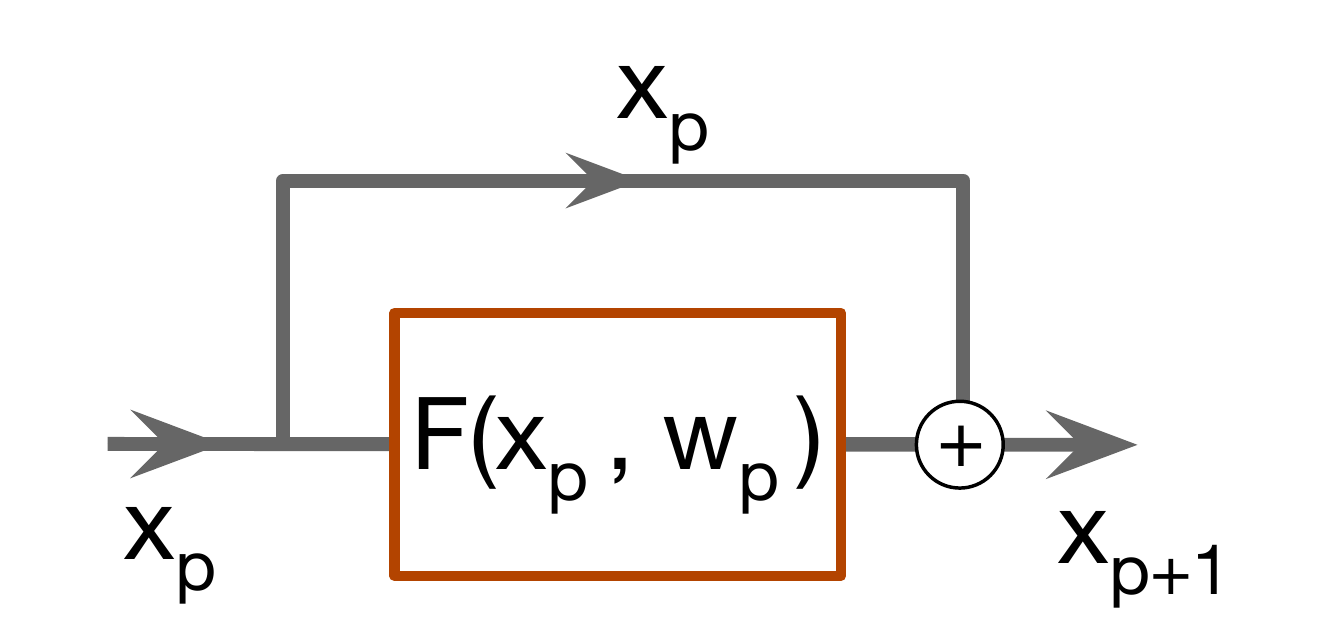}
  \caption{A block with residual connections.}
  \vspace{-10pt}
  \label{res_block}
\end{wrapfigure}
Residual connections were first introduced and later refined by He et
al.~\cite{he2015deep,he2016identity} for the effective training of deep
networks.  The key idea of residual connection is to create identity mapping
connections to bypass the parameterised layers in a network. The input of a
residual block is directly merged to the output by addition.  The residual
connections have been shown to make information propagation smooth and improve
the training speed~\cite{he2015deep}.

More specifically, let the input to the $p$-th layer of a residual block as
$\mathbf{x}_p$, the output of the block $\mathbf{x}_{p+1}$ has the form:
$\mathbf{x}_{p+1} = \mathbf{x}_{p} + F(\mathbf{x}_{p}, \mathbf{w}_{p})$;
where $F(\mathbf{x}_{p},\mathbf{w}_{p})$ denotes the path with non-linear
functions in the block (shown in Fig.~\ref{res_block}).
If we stack the residual blocks, the last layer output $\mathbf{x}_{l}$
can be expressed as:
$\mathbf{x}_{l} = \mathbf{x}_{p} +
\sum_{i=p}^{l-1} F(\mathbf{x}_{i}, \mathbf{w}_{i})$.
The residual connections enables direct information propagation from
any residual block to another in both forward pass and back-propagation.


\subsubsection{Effective receptive field.}
One interpretation of the residual network is that they behave like ensembles
of relatively shallow networks.  The unravelled view of
the residual connections proposed by Veit et al.~\cite{veit2016residual}
suggests that the networks with $n$ residual blocks have a collection of $2^n$
unique paths.

\begin{wrapfigure}{r}{0.40\textwidth}
  \centering
  \vspace{-20pt}
  \includegraphics[width=1.0\linewidth]{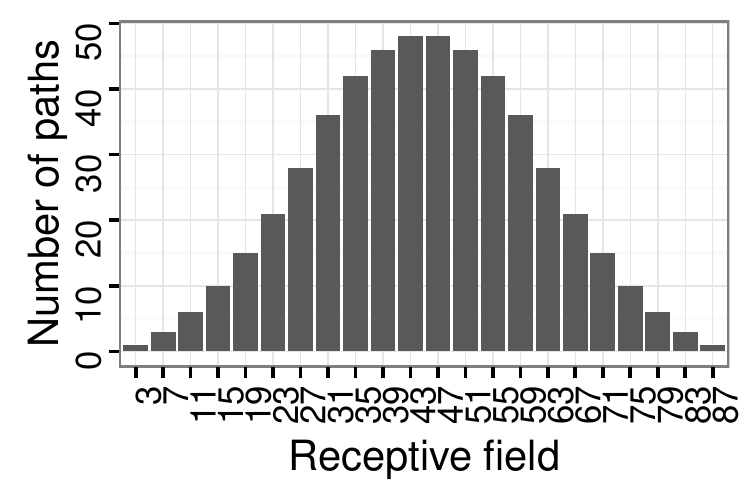}
  \vspace{-20pt}
  \caption{Histogram of the receptive fields.}
  \vspace{-20pt}
  \label{receptive_field}
\end{wrapfigure}

Without residual connections, the receptive field of a network is generally
considered fixed.  However, when training with $n$ residual blocks, the
networks utilise $2^n$ different paths and therefore features can be
learned with a large range of different receptive fields.  For example, the
proposed network with $9$ residual blocks (see Section~\ref{proposednetwork})
has a maximum receptive field of $87\times87\times87$ voxels.  Following the
unravel view of the residual network, it consists of $2^9$ unique paths.
Fig.~\ref{receptive_field} shows the distribution of the receptive field of
these paths.  The receptive fields range from $3\times 3\times 3$ to $87\times
87\times 87$, following a binomial distribution.

This differs from the existing 3D networks.  For example,
Deepmedic~\cite{Kamnitsas201761} model operates at two paths, with a fixed
receptive field $17\times 17\times 17$ and $42\times 42\times 42$ respectively.
3D U-net~\cite{cciccek20163d} has a relatively large receptive field of about
$88\times 88\times 88$ voxels. However, there are only eight unique paths and
receptive fields.

Intuitively, given that the receptive field of a deep convolutional network is
relatively large, the segmentation maps will suffer from distortions due to the
border effects of convolution.  That is, the segmentation results near the
border of the output volume are less accurate due to the lack of input
supporting window.  We conduct experiments and demonstrate that the proposed
networks generate only a small distortion near the borders (See
Section~\ref{rf_exp}).  This suggests training the network with residual
connections reduces the effective receptive field.  The width of the distorted
border is much smaller than the maximum receptive field.  This phenomenon was
also recently analysed by Luo et al.~\cite{luo2016understanding}.  In practice,
at test time we pad each input volume with a border of zeros and discard the
same amount of border in the segmentation output.

\subsubsection{Loss function.}
The last layer of the network is a softmax function that gives scores over all
labels for each voxel.  Typically, the end-to-end training procedure minimises
the cross entropy loss function using an $N$-voxel image volume
$\{x_n\}_{n=1}^{N}$ and the training data of $C$-class segmentation map
$\{y_n\}_{n=1}^{N}$ where $y_n\in \{1,\dots,C\}$ is:
\begin{align}
  \mathcal{L}(\{x_n\}, \{y_n\}) =
  - \frac{1}{N}\sum_{n=1}^{N}\sum_{c=1}^{C}\delta(y_n=c)\log F_c(x_n),
  \label{crossentropy_equation}
\end{align}
where $\delta$ corresponds to the Dirac delta function, $F_c(x_n)$ is the
softmax classification score of $x_n$ over the $c$-th class.  However, when the
training data are severely unbalanced (which is typical in medical image
segmentation problems), this formulation leads to a strongly biased estimation
towards the majority class.  Instead of directly re-weighting each voxel by
class frequencies, Milletari et al.~\cite{milletari2016v} propose a solution by
maximising the mean Dice coefficient directly, i.e.,
\begin{align}
  \mathcal{D}(\{x_n\}, \{y_n\}) =
  \frac{1}{C}\sum_{c=1}^C \frac{2\sum_{n=1}^N \delta(y_n=c)F_c(x_n)}
  {\sum_{n=1}^N [\delta(y_n=c)]^2 + \sum_{n=1}^N [F_c(x_n)]^2}.
\end{align}
We employ this formulation to handle the issue of training data imbalance.

\subsubsection{Uncertainty estimation using dropout.}
Gal and Ghahramani demonstrated that the deep network trained with dropout can
be cast as a Bayesian approximation of the Gaussian
process~\cite{Gal2016Dropout}.  Given a set of training data and their labels
$\{\mathbf{X}, \mathbf{Y}\}$, training a network $F(\cdot\,,\mathbf{W})$
with dropout has the effect of approximating the posterior distribution
$p(\mathbf{W} | \{\mathbf{X}, \mathbf{Y}\})$ by minimising the Kullback-Leibler
divergence term, i.e. $\textrm{KL}(q(\mathbf{W})||p(\mathbf{W} | \{\mathbf{X},
\mathbf{Y}\}))$; where $q(\mathbf{W})$ is an approximating distribution over
the weight matrices $\mathbf{W}$ with their elements randomly set to zero
according to Bernoulli random variables.
After training the network, the predictive distribution of test data
$\mathbf{\hat{x}}$ can be expressed as $q(\mathbf{\hat{y}}|\mathbf{\hat{x}}) =
\int F(\mathbf{\hat{x}},\mathbf{W})q(\mathbf{W})d\mathbf{W}$.  The prediction
can be approximated using Monte Carlo samples of the trained network:
$\mathbf{\hat{y}} = \frac{1}{M} \sum_{m=1}^{M} F(\mathbf{\hat{x}},
\mathbf{W}_m)$, where $\{\mathbf{W}_m\}_{m=1}^{M}$ is a set of $M$ samples from
$q(\mathbf{W})$.  The uncertainty of the prediction can be estimated using the
sample variance of the $M$ samples.

With this theoretical insight, we are able to estimate the uncertainty of the
segmentation map at the voxel level.  We extend the segmentation network with a
$1\times 1\times 1$ convolutional layer before the last convolutional layer.
The extended network is trained with a dropout ratio of 0.5 applied to the
newly inserted layer. At test time, we sample the network N times using
dropout. The final segmentation is obtained by majority voting. The percentage
of samples which disagrees with the voting results is calculated at each voxel
as the uncertainty estimates.


\section{The network architecture and its implementation}
\label{proposednetwork}
\subsection{The proposed architecture}
\begin{figure}[t]
  \includegraphics[width=1.0\linewidth]{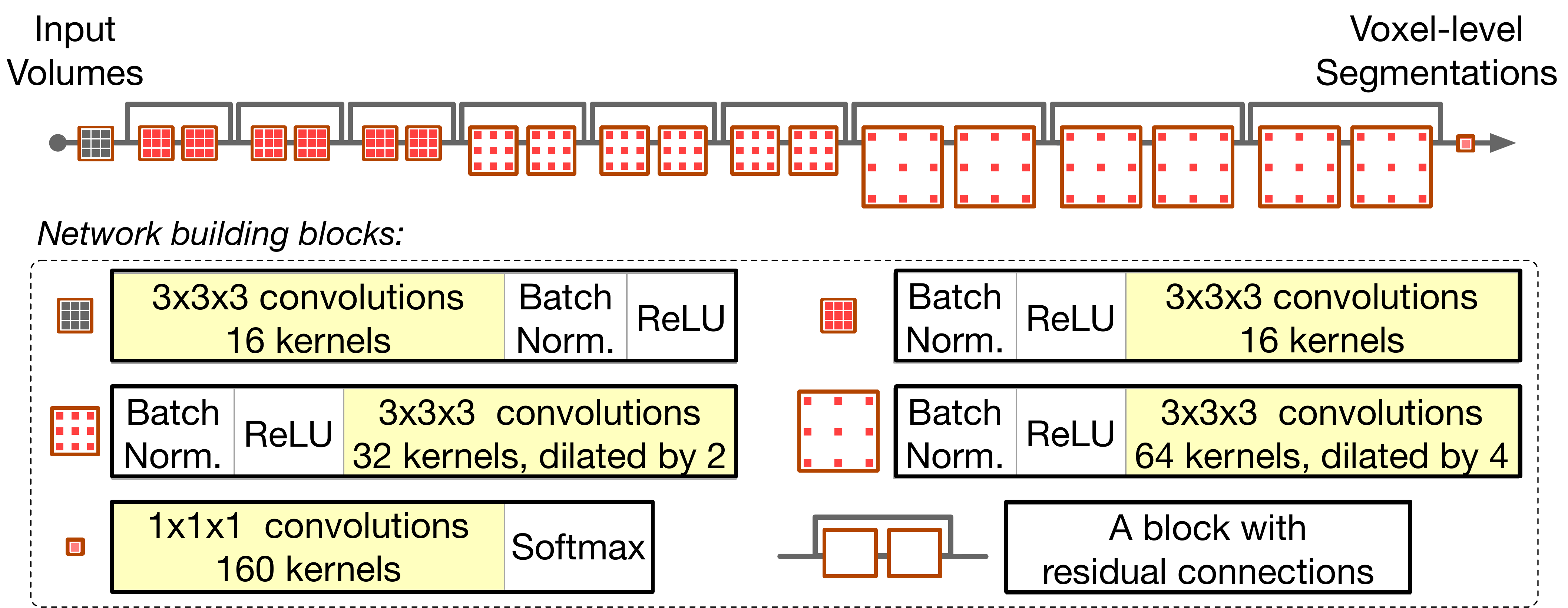}
  \caption{The proposed network architecture for volumetric image segmentation.
    The network mainly utilises dilated convolutions and residual connections
    to make an end-to-end mapping from image volume to a voxel-level dense
    segmentation.  To incorporate features at multiple scales, the dilation
    factor of the dilated convolutions is gradually increased when the layer
    goes deeper.  The residual blocks with identity mapping enable the direct
    fusion of features from different scales.  The spatial resolution of the
    input volume is maintained throughout the network.}
  \label{basic_net}
\end{figure}
Our network consists of $20$ layers of convolutions.   In the first seven
convolutional layers, we adopt $3\times3\times3$-voxel convolutions.  These
layers are designed to capture low-level image features such as edges and
corners.  In the subsequent convolutional layers, the kernels are dilated by a
factor of two or four.  These deeper layers with dilated kernels encode mid-
and high-level image features.

Residual connections are employed to group every two convolutional layers.
Within each residual block, each convolutional layer is associated with an
element-wise rectified linear unit (ReLU) layer and a batch normalisation
layer~\cite{ioffe2015batch}.  The ReLU, batch normalisation, and convolutional
layers are arranged in the pre-activation order~\cite{he2016identity}.

The network can be trained end-to-end.  In the training stage, the inputs to our
network are $96\times96\times96$-voxel images.  The final softmax layer gives
classification scores over the class labels for each of the
$96\times96\times96$ voxels.  The architecture is illustrated in
Fig.~\ref{basic_net}.

\subsection{Implementation details}
In the training stage, the pre-processing step involved input data
standardisation and augmentation at both image- and subvolume-level.  At
image-level, we adopted the histogram-based scale standardisation
method~\cite{nyul2000new} to normalised the intensity histograms.  As a data
augmentation at image-level, randomisation was introduced in the
normalisation process by randomly choosing a threshold of foreground between
the volume minimum and mean intensity (at test time, the mean intensity
of the test volume was used as the threshold).  Each image was
further normalised to have zero mean and unit standard deviation.  Augmentations
on the randomly sampled $96\times 96\times 96$ subvolumes were employed on the
fly.  These included rotation with a random angle in the range of $[-10\degree,
10\degree]$ for each of the three orthogonal planes and spatial rescaling
with a random scaling factor in the range of $[0.9, 1.1]$.

All the parameters in the convolutional layers were initialised according to He
et al.~\cite{he2015delving}.  The scaling and shifting parameters in the batch
normalisation layers were initialised to $1$ and $0$ respectively.  The
networks were trained with two Nvidia K80 GPUs.  At each training iteration,
each GPU processed one input volume; the average gradients computed over these
two training volumes were used as the gradients update.  To make a fair
comparison, we employed the Adam optimisation method~\cite{kingma2014adam} for
all the methods with fixed hyper-parameters.  The learning rate $lr$ was set to
$0.01$, the step size hyper-parameter $\beta_1$ was $0.9$ and $\beta_2$ was
$0.999$ in all cases, except V-Net for which we chose the largest $lr$ that the
training algorithm converges ($lr=0.0001$).  The models were trained until we
observed a plateau in performance on the validation set.  We do not employ
additional spatial smoothing function (such as conditional random field) as a
post-processing step.  Instead of aiming for better segmentation results by
adding post-processing steps, we focused on the dense segmentation maps
generated by the networks.  As we consider brain parcellation as a pretext
task, networks without explicit spatial smoothing are potentially more
reusable.  We implemented all the methods (including a re-implementation of
Deepmedic~\cite{Kamnitsas201761}, V-net~\cite{milletari2016v}, and 3D
U-net~\cite{cciccek20163d} architecture) with Tensorflow\footnote{ The source
  code is available at \url{https://github.com/gift-surg/HighRes3DNet}}.

\section{Experiments and results}
\subsubsection{Data.}
To demonstrate the feasibility of learning complex 3D image representations
from large-scale data, the proposed network is learning a highly granular
segmentation of $543$ T1-weighted MR images of healthy controls from the ADNI
dataset.  The average number of voxels of each volume is about $182\times
244\times 246$.  The average voxel size is approximately $1.18mm\times
1.05mm\times 1.05mm$.  All volumes are bias-corrected and reoriented to a
standard Right-Anterior-Superior orientation.  The bronze standard parcellation
of $155$ brain structures and $5$ non-brain outer tissues are obtained using
the GIF framework~\cite{cardoso2015geodesic}.  Fig.~\ref{overall_boxplot}(left)
shows the label distribution of the dataset.  We randomly choose 443, 50, and
50 volumes for training, test, and validation respectively.

\subsubsection{Overall evaluation.}
\begin{figure}[t]
  \centering
  \begin{minipage}{.19\textwidth}
  \centering
    \includegraphics[width=1.0\linewidth]{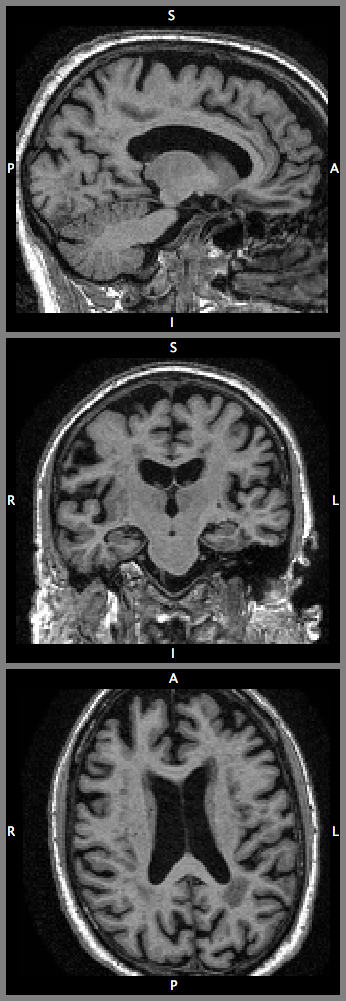}\\
    (1)
  \end{minipage}
  \begin{minipage}{.19\textwidth}
  \centering
    \includegraphics[width=1.0\linewidth]{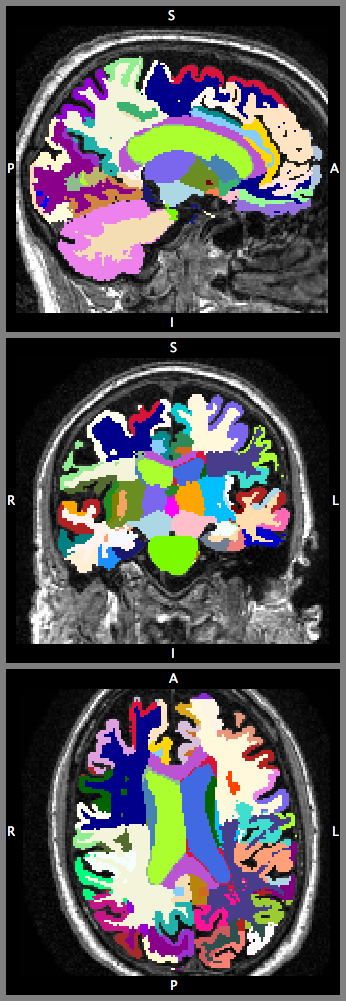}\\
    (2)
  \end{minipage}
  \begin{minipage}{.19\textwidth}
  \centering
    \includegraphics[width=1.0\linewidth]{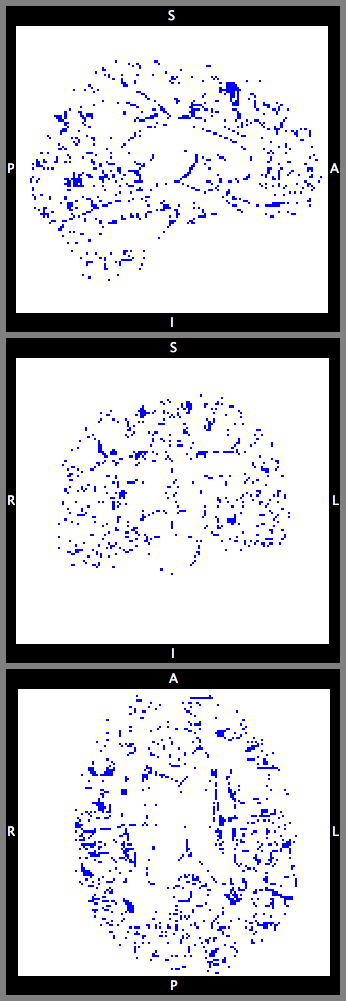}\\
    (3)
  \end{minipage}
  \begin{minipage}{.19\textwidth}
  \centering
    \includegraphics[width=1.0\linewidth]{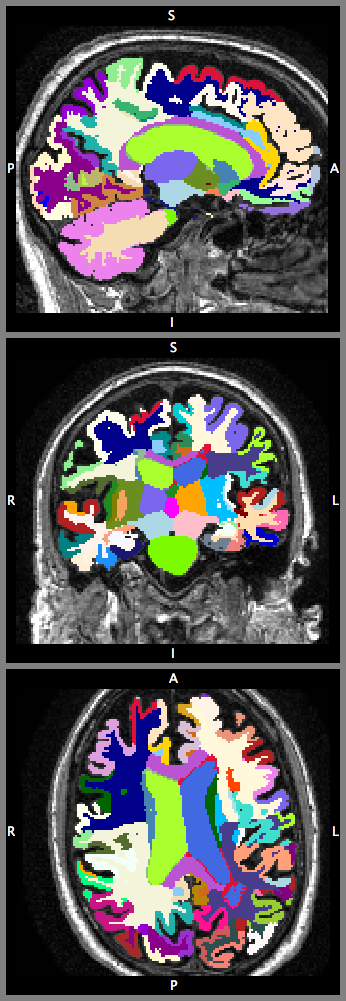}\\
    (4)
  \end{minipage}
  \begin{minipage}{.19\textwidth}
  \centering
    \includegraphics[width=1.0\linewidth]{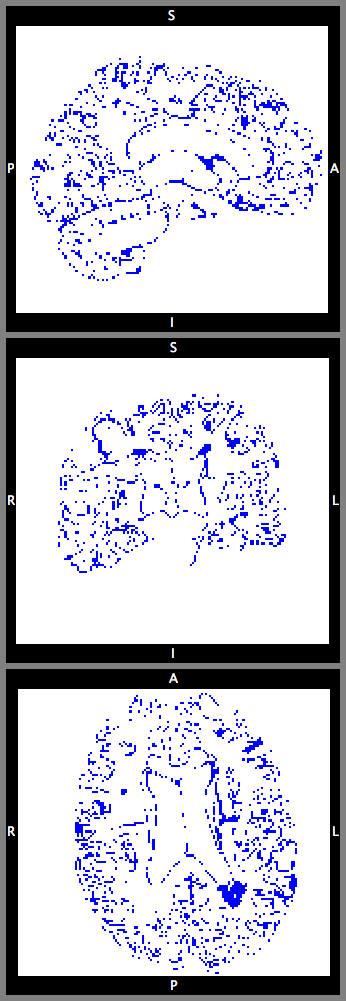}\\
    (5)
  \end{minipage}
  \caption{Visualisations of segmentation results.  (1) slices from a test
    image volume, segmentation maps and false prediction maps
    generated by HC-dropout (2, 3), and 3D U-net-dice (4, 5).  }
  \label{example}
\end{figure}
\begin{figure}[tb]
  \centering
  \begin{minipage}{.49\textwidth}
    \centering
    \includegraphics[width=1.0\linewidth]{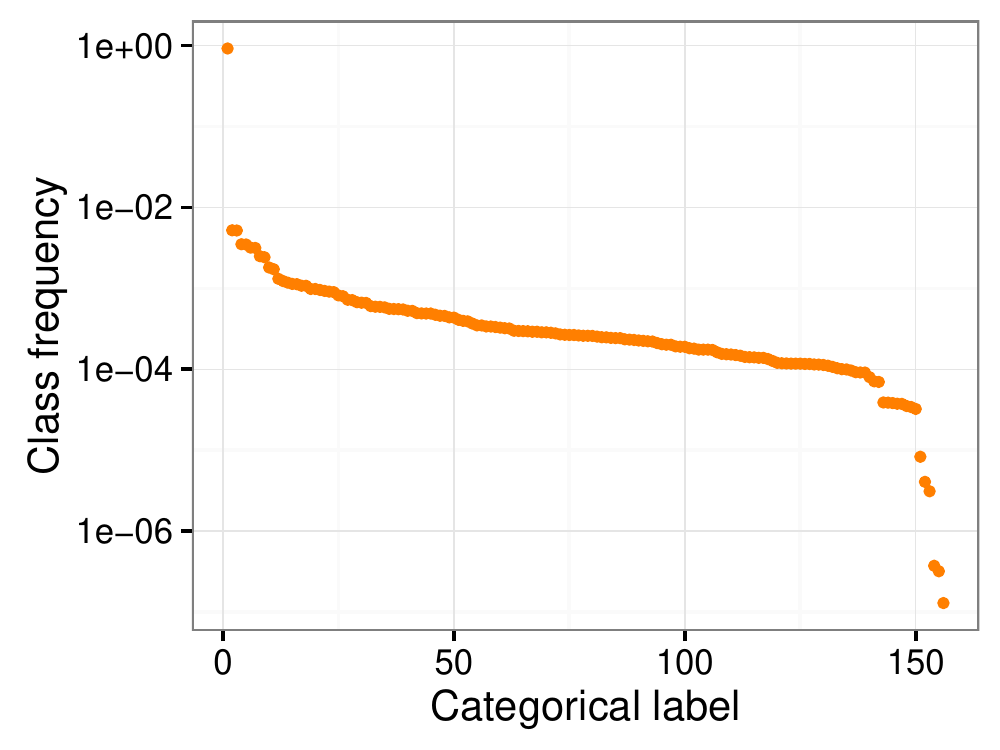}
  \end{minipage}
  \begin{minipage}{.49\textwidth}
    \centering
    \includegraphics[width=1.0\linewidth]{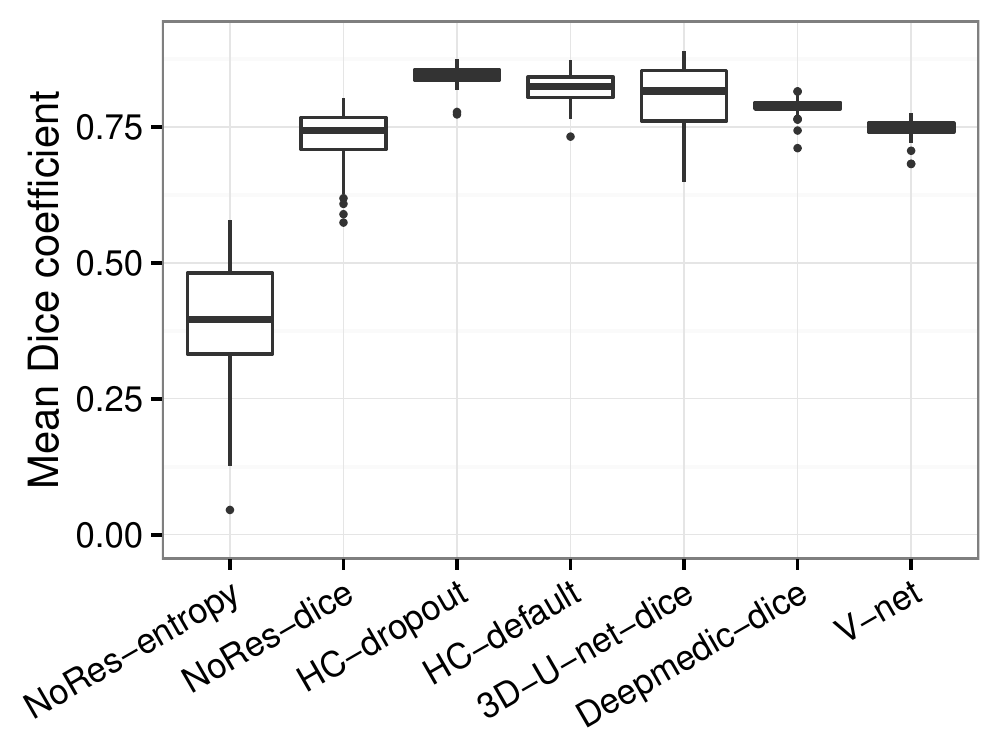}
  \end{minipage}
  \caption{Left: label distribution of the dataset;  right: comparison of
  different network architectures.  }
  \label{overall_boxplot}
\end{figure}
\begin{table}[t]
  \centering
  \caption{Comparison of different 3D convolutional network architectures.}
\begin{tabular}{lccccc}
  \toprule
	Architecture & Multi-layer fusion & Num. param. & Loss type & DCS (\%) & STD (\%) \\
  \midrule
	HC-default & Residual & 0.81M & Dice loss & 82.05 & 2.96  \\
  HC-dropout & Residual & 0.82M & Dice loss & \textbf{84.34} & \textbf{1.89}  \\
	NoRes-entropy & N/A & 0.81M & Cross entr. & 39.36 & 1.13  \\
	NoRes-dice & N/A & 0.81M & Dice loss  & 75.47 & 2.97  \\
  Deepmedic\cite{Kamnitsas201761}-dice& Two pathways & 0.68M & Dice loss &  78.74  & 1.72  \\
	3D U-net\cite{cciccek20163d}-dice& Feature forwarding & 19.08M & Dice loss & 80.18  & 6.18  \\
	V-net\cite{milletari2016v} & Feature forwarding & 62.63M & Dice loss & 74.58  &  1.86  \\
  \bottomrule
  \label{overall_cmp}
  \vspace{-20pt}
\end{tabular}
\end{table}
\begin{figure}
  \includegraphics[width=1.0\linewidth]{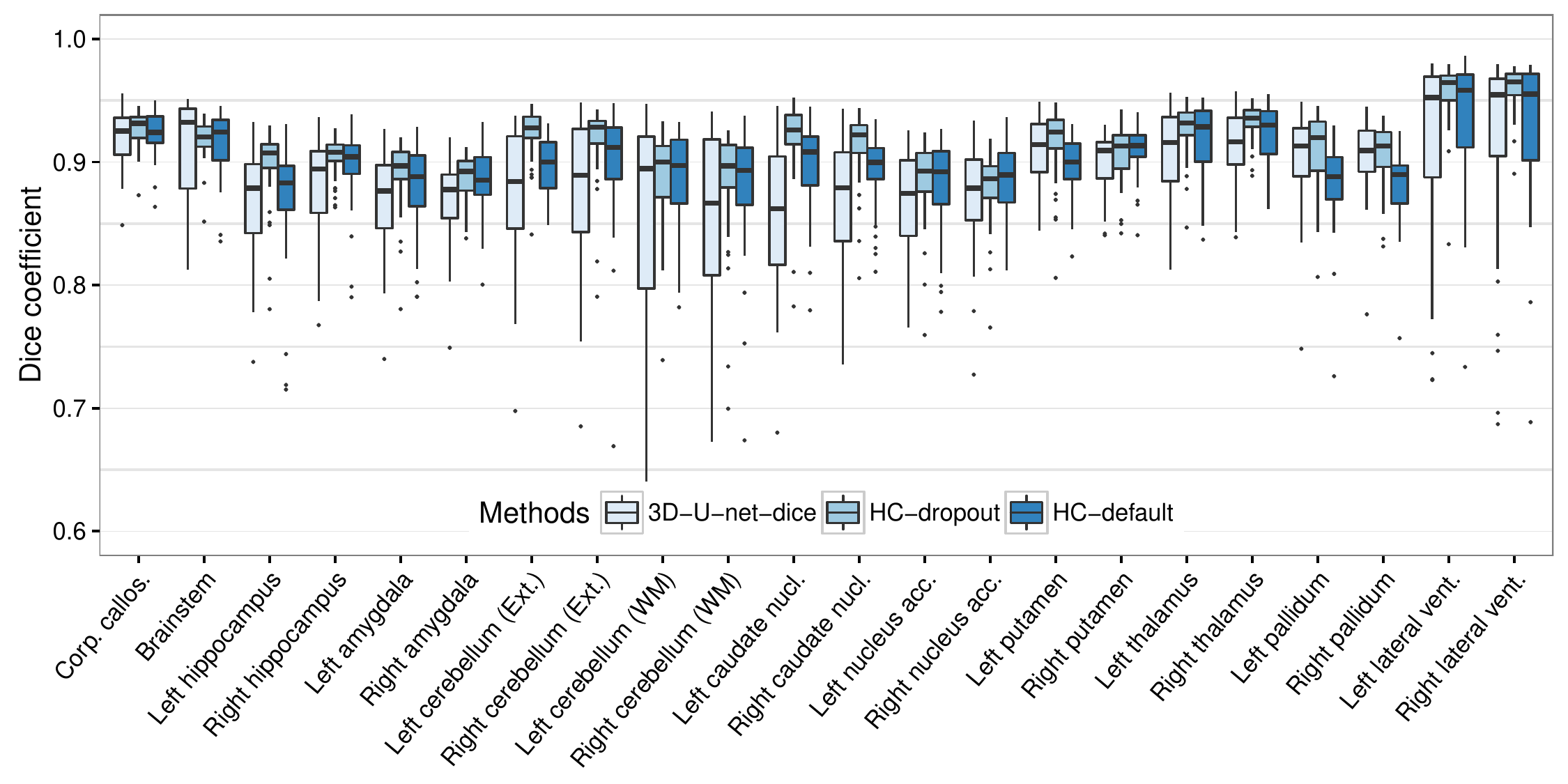}
  \caption{Segmentation performance against a set of key structures.}
  \label{per_class}
\end{figure}

In this section, we compare the proposed high-resolution compact network
architecture (illustrated in Fig.~\ref{basic_net}; denoted as {\it HC-default})
with three variants: (1) the HC-default configuration without the residual
connections, trained with cross-entropy loss function ({\it NoRes-entropy});
(2) the HC-default configuration without residual connections, trained with
Dice loss function ({\it NoRes-dice}); and (3) the HC-default configuration
trained with an additional dropout layer, and makes predictions with a majority
voting of $10$ Monte Carlo samples ({\it HC-dropout}).  For the dropout
variant, our dropout layer employed before the last convolutional layer
consists of $80$ kernels.

Additionally, three state-of-the-art volumetric segmentation networks are
evaluated.  These include 3D U-net~\cite{cciccek20163d},
V-net~\cite{milletari2016v}, and Deepmedic~\cite{Kamnitsas201761}.  The last
layer of each network architecture is replaced with a 160-way softmax
classifier.

We observe that training these networks with the cross entropy loss function
(Eq.~\ref{crossentropy_equation}) leads to poor segmentation results.  Since
the cross-entropy loss function treats all training voxels equally, the network
may have difficulties in learning representations related to the minority
classes.  Training with the Dice loss function alleviates this issue by
implicitly re-weighting the voxels.  Thus we train all networks using the Dice
loss function for a fair comparison.

We use the mean Dice Coefficient Similarity (DCS) as the performance measure.
Table~\ref{overall_cmp} and Fig.~\ref{overall_boxplot}(right) compare the
performance on the test set.  In terms of our network variants, the results
show that the use of Dice loss function largely improves the segmentation
performance.  This suggests that the Dice loss function can handle the severely
unbalanced segmentation problem well.  The results also suggest that
introducing the residual connections improved the segmentation performance
measured in mean DCS.  This indicates that the residual connections are
important elements of the proposed network.  By adopting the dropout method,
the DCS can be further improved by 2\% in DCS.

With a relatively small number of parameters, our HC-default and HC-dropout
outperform the competing methods in terms of mean DCS.  This suggests
that our network is more effective for the brain parcellation problem.  Note
that V-net has a similar architecture to 3D U-net and has more parameters,
but does not employ the batch normalisation technique. The lower DCS produced
by V-net suggests that batch normalisation is important for training the
networks for brain parcellation.

In Fig.~\ref{per_class}, we show that the dropout variant achieves better
segmentation results for all the key structures.  Fig.~\ref{example} presents
an example of the segmentation results of the proposed network and 3D
U-net-Dice.

\subsubsection{Receptive field and border effects.}
\label{rf_exp}
We further compare the segmentation performance of a trained network by
discarding the borders in each dimension of the segmentation map.  That is,
given a $d\times d\times d$-voxel input,  at border size $1$ we only preserve the
$(d-2)^3$-voxel output volume centred within the predicted map.
\begin{wrapfigure}{r}{0.5\linewidth}
  \centering
  \includegraphics[width=0.8\linewidth]{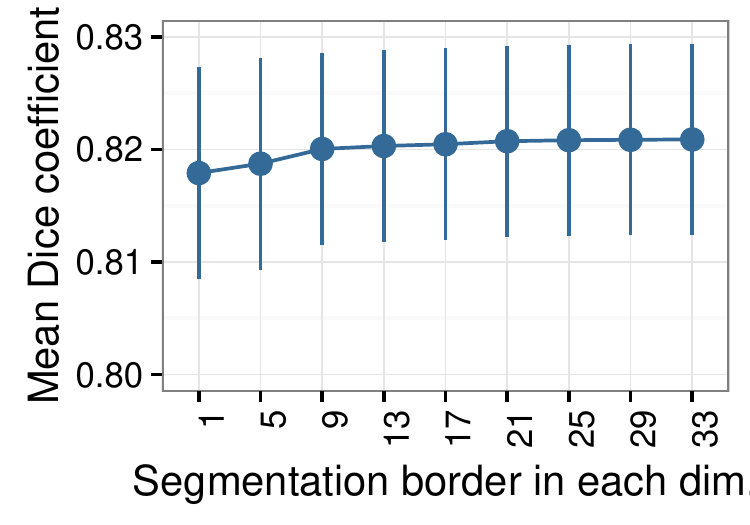}\\
  \vspace{-10pt}
  \caption{Empirical analysis of the segmentation borders.
  Voxels near to the volume borders are classified less accurately.}
  \vspace{-30pt}
  \label{window_boxplot}
\end{wrapfigure}
Fig.~\ref{window_boxplot} plots the DCS and standard errors of segmentation
according to the size of the segmentation borders in each dimension.  The
results show that the distorted border is around $17$ voxels in each dimension.
The border effects do not severely decrease the segmentation performance.  In
practice, we pad the volume images with $16$ zeros in each dimension, and remove
the same amount of borders in the segmentation output.

\subsubsection{The effect of number of samples in uncertainty estimations.}
\begin{figure}[b]
  \centering
  \vspace{-15pt}
  \begin{minipage}{.48\textwidth}
    \centering
    \includegraphics[width=1.0\linewidth]{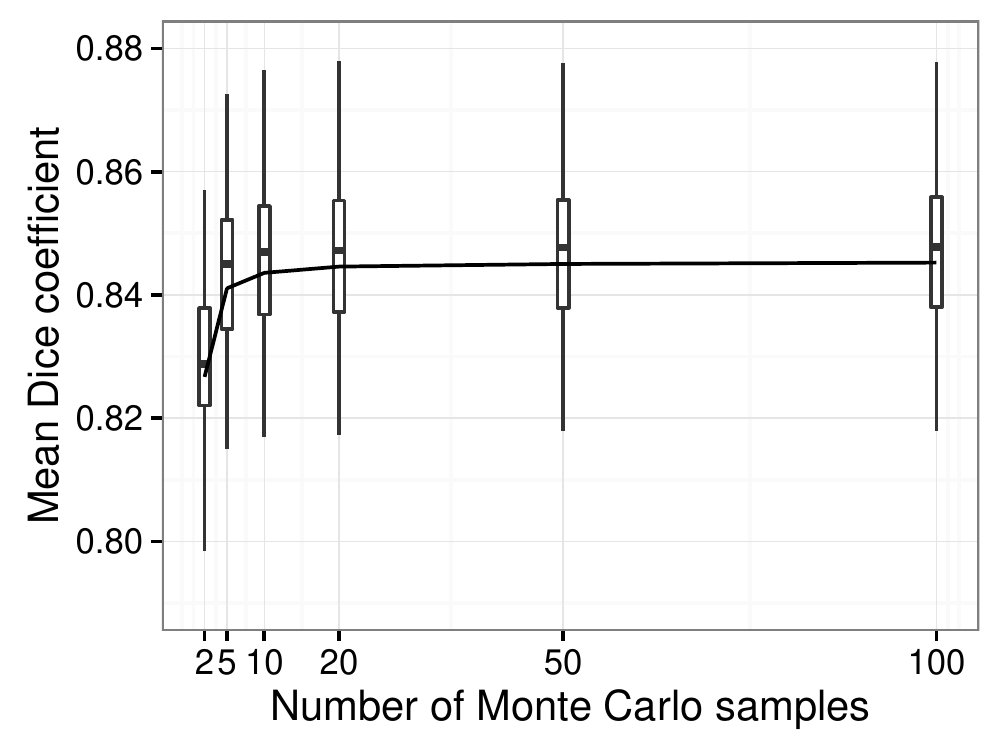}\\
    (a)
  \end{minipage}
  \begin{minipage}{.48\textwidth}
    \centering
    \includegraphics[width=0.75\linewidth]{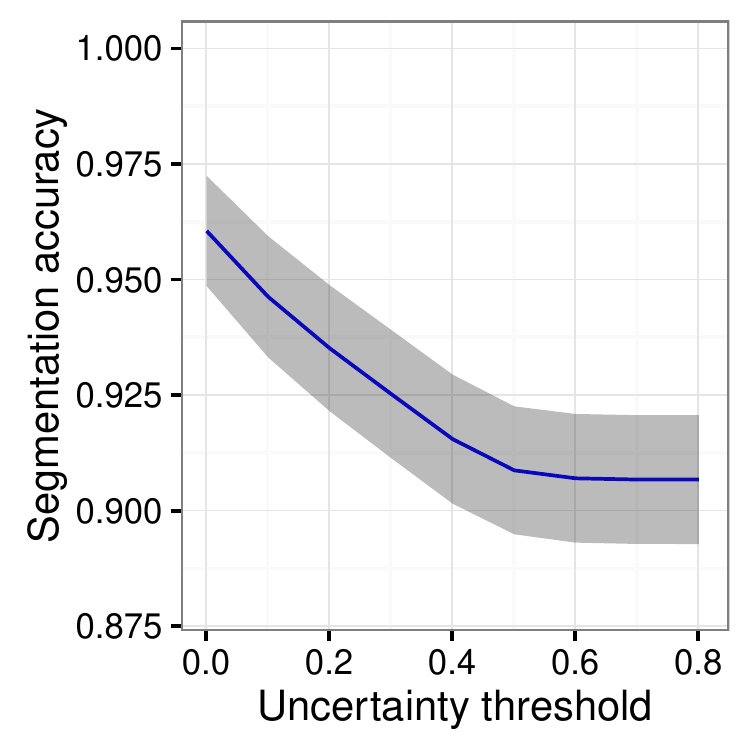}\\
    (b)
  \end{minipage}
  \caption{Evaluation of dropout sampling.  (a) The segmentation performance
    against the number of Monte Carlo samples.  (b) voxel-level segmentation
    accuracy by thresholding the uncertainties. The shaded area represents
    the standard errors.}
  \label{n_threshold}
\end{figure}
\begin{figure}
  \centering
  \includegraphics[width=0.30\linewidth]{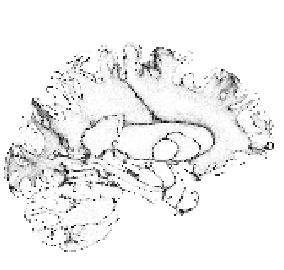}
  \includegraphics[width=0.30\linewidth]{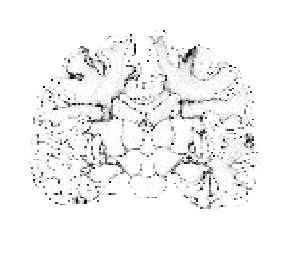}
  \includegraphics[width=0.30\linewidth]{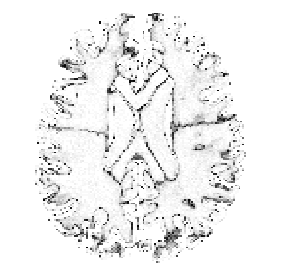}
  \smallskip
  \includegraphics[width=0.30\linewidth]{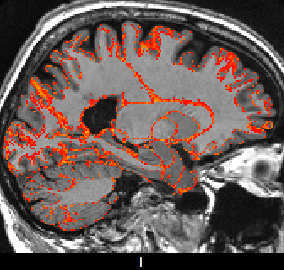}
  \includegraphics[width=0.30\linewidth]{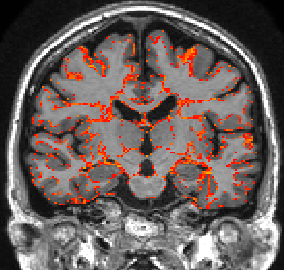}
  \includegraphics[width=0.30\linewidth]{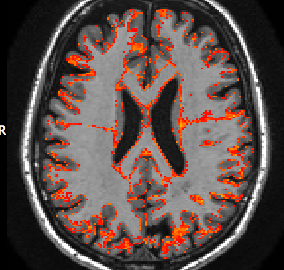}
  \caption{Voxel-level segmentation uncertainty estimations.
  Top row: uncertainty map generated with $100$ Monte Carlo samples using dropout.
  Bottom row: uncertainty map thresholded at $0.1$.}
  \label{uncertainty}
\end{figure}
This section investigates the number of Monte Carlo samples and the segmentation
performance of the proposed network.  Fig.~\ref{n_threshold}(a) suggests that using
$10$ samples is enough to achieve good segmentation.  Further increasing the
number of samples has relatively small effects on the DCS.
Fig.~\ref{n_threshold}(b) plots the voxel-wise segmentation accuracy computed
using only the voxels with an uncertainty less than a threshold.
The voxel-wise accuracy is high when the threshold is small.
This indicates that the uncertainty estimation reflects the
confidence of the network.
Fig.~\ref{uncertainty} shows an uncertainty map generated by the proposed
network.  The uncertainties near the boundaries of different
structures are relatively higher than the other regions.

Currently, our method takes about $60$ seconds to predict a typical volume with
$192\times 256\times 256$ voxels.  To achieve better segmentation results and
measure uncertainty, $10$ Monte Carlo samples of our dropout model are
required.  The entire process takes slightly more than $10$ minutes in total.
However, during the Monte Carlo sampling at test time, only the dropout layer
and the final prediction layer are randomised.  To further reduce the
computational time, the future software could reuse the features extracted from
the layers before dropout, resulting in only a marginal increase in runtime
when compared to a single prediction.

\section{Conclusion}
In this paper, we propose a high-resolution, 3D convolutional network
architecture that incorporates large volumetric context using dilated
convolutions and residual connections.  Our network is conceptually simpler and
more compact than the state-of-the-art volumetric segmentation networks.
We validate the proposed network using the challenging task of brain
parcellation in MR images.  We show that the segmentation performance of our
network compares favourably with the competing methods.  Additionally, we
demonstrate that Monte Carlo sampling of dropout technique can be used to
generate voxel-level uncertainty estimation for our brain parcellation network.
Moreover, we consider the brain parcellation task as a pretext task for
volumetric image segmentation.  Our trained network potentially provides a good
starting point for transfer learning of other segmentation tasks.

In the future, we will extensively test the generalisation ability of the
network to brain MR scans obtained with various scanning protocols from
different data centres.  Furthermore, we note that the uncertainty estimations
are not probabilities.  We will investigate the calibration of the uncertainty
scores to provide reliable probability estimations.

\subsubsection*{Acknowledgements.}
This work was supported through an Innovative Engineering for Health award by
the Wellcome Trust [WT101957, 203145Z/16/Z], Engineering and Physical Sciences
Research Council (EPSRC) [NS/A000027/1, NS/A000050/1], the National Institute
for Health Research University College London Hospitals Biomedical Research
Centre (NIHR BRC UCLH/UCL High Impact Initiative), UCL EPSRC CDT Scholarship
Award [EP/L016478/1], a UCL Overseas Research Scholarship, a UCL Graduate
Research Scholarship, and the Health Innovation Challenge Fund [HICF-T4-275, WT
97914], a parallel funding partnership between the Department of Health and
Wellcome Trust.  The authors would also like to acknowledge that the work
presented here made use of Emerald, a GPU-accelerated High Performance
Computer, made available by the Science \& Engineering South Consortium
operated in partnership with the STFC Rutherford-Appleton Laboratory.

\bibliographystyle{splncs03}
\bibliography{bib_brain_seg}
\end{document}